\documentclass[conference]{IEEEtran}
\IEEEoverridecommandlockouts
\usepackage{cite}
\usepackage{amsmath,amssymb,amsfonts}
\usepackage{algorithmic}
\usepackage{graphicx}
\usepackage{textcomp}
\usepackage{booktabs}
\usepackage{xcolor}
\usepackage{multirow}
\usepackage{makecell}
\usepackage{listings}
\usepackage[frozencache,cachedir=.]{minted}
\usepackage{tcolorbox}  
\usepackage{tablefootnote}
\usepackage{threeparttable}
\usepackage{enumitem}
\usepackage{hyperref}
\usepackage{xcolor}
\hypersetup{
    colorlinks,
    citecolor={green!50!black},
}

\def\BibTeX{{\rm B\kern-.05em{\sc i\kern-.025em b}\kern-.08em
    T\kern-.1667em\lower.7ex\hbox{E}\kern-.125emX}}

\begin{document}

\title{KirchhoffNet: A Scalable Ultra Fast Analog Neural Network
}

\author{
Zhengqi Gao$^1$, Fan-Keng Sun$^{1}$, Ron Rohrer$^2$, Duane S. Boning$^1$\\ 
$^1$ Department of EECS, Massachusetts Institute of Technology, Cambridge, MA 02139, USA
\\
$^{2}$ Department of ECE, Carnegie Mellon University, Pittsburgh, PA 15213, USA
\\
}

\maketitle

\begin{abstract}
In this paper, we leverage a foundational principle of analog electronic circuitry, Kirchhoff's current and voltage laws, to introduce a distinctive class of neural network models termed \emph{KirchhoffNet}. Essentially, KirchhoffNet is an analog circuit that can function as a neural network, utilizing its initial node voltages as the neural network input and the node voltages at a specific time point as the output. The evolution of node voltages within the specified time is dictated by learnable parameters on the edges connecting nodes. We demonstrate that KirchhoffNet is governed by a set of ordinary differential equations (ODEs), and notably, even in the absence of traditional layers (such as convolution layers), it attains state-of-the-art performances across diverse and complex machine learning tasks. Most importantly, KirchhoffNet can be potentially implemented as a low-power analog integrated circuit, leading to an appealing property --- irrespective of the number of parameters within a KirchhoffNet, its on-chip forward calculation can always be completed within a short time. This characteristic makes KirchhoffNet a promising and fundamental paradigm for implementing large-scale neural networks, opening a new avenue in analog neural networks for AI.
\end{abstract}


\section{Introduction}\label{sec:intro}

Despite the availability of several alternative platforms such as CPUs, edge devices~\cite{chen2019deep_edge}, and FPGAs~\cite{wang2016dlau}, GPUs stand out as the foremost choice for deploying contemporary deep neural networks (DNNs)~\cite{oh2004gpu, owens2008gpu, nickolls2010gpu}. The acceleration of DNN execution on GPUs is achieved through extensive parallelization and optimization of computational operations~\cite{nickolls2010gpu, oh2004gpu, owens2008gpu}, notably matrix-vector multiplications (MVMs). To date, even though GPUs have undergone continuous improvement, they barely manage to keep up with the escalating scale of neural networks. Moreover, with Moore's law nearing its limits, GPUs are likely to fall behind in the race, as continually doubling transistor density into the foreseeable future appears to be unlikely. In addition, the substantial power consumption of GPUs has become a major concern. These factors, coupled with the pursuit of even faster computing paradigms, have prompted researchers to explore alternative approaches.

Several emerging technologies have garnered significant attention~\cite{wiebe2014quantum,li2015merging,shen2017deep,burr2017neuromorphic,biamonte2017quantum,schuman2017survey_neuromorphic,potok2018study_neuromorphic,lin2018all_optical,xia2019memristive,eleftheriou2019deep_imc,bouvier2019spiking,rekhi2019analog,yin2020imc,sebastian2020memory,garg2020advances_quantum,roy2020memory,yu2021compute_imc,dillavou2022demonstration}. Some are generic in the sense that they aim to accelerate MVM operations, enabling the efficient execution of various types of DNNs, while others propose hardware designed specifically for particular types of neural networks. To name a few examples, a photonic Mach–Zehnder interferometer can realize MVM in the optical domain~\cite{shen2017deep}, which potentially portends lower power consumption and greater bandwidth in comparison with electronic-based computation. In-memory computing~\cite{roy2020memory,yu2021compute_imc} suggests using memory elements not only to store information but also to perform MVM operations~\cite{xia2019memristive}, thereby saving data accessing time. Neuromorphic computing, inspired by the working of the human brain, manipulates time-domain spikes and is well-suited for realizing spiking neural networks~\cite{schuman2017survey_neuromorphic}. Efforts have also been made using quantum computing devices to carry out quantum machine learning~\cite{wiebe2014quantum,biamonte2017quantum,garg2020advances_quantum,jiang2021machine_quantum}. Other notable approaches include but are not limited to reservoir computing~\cite{gallicchio2017deep_reservior}, ferroelectric neural network~\cite{jerry2017ferroelectric}, stochastic computing~\cite{ren2017sc}, and spintronics-based computing~\cite{jia2018spintronics}.

In this paper, we propose a new class of neural network models by exploiting Kirchhoff's current and voltage laws, and thus we coin the name~\emph{KirchhoffNet}. A KirchhoffNet resembles an analog circuit, with the initial node voltages at $t=0$ being the neural network inputs, and the node voltages at $t=T$ being read out as the neural network outputs. Learnable parameters on the edges connecting nodes influence branch currents and dictate the node voltage evolution within the time interval $[0,T]$ through a set of ordinary differential equations (ODEs). The adjoint method~\cite{director1969generalized,li2023adjoint,chen2018neural} can be used to calculate the derivatives of a customized loss function with respect to the learnable parameters, enabling the use of gradient-based optimizers, such as SGD and Adam, for KirchhoffNet training.

Notably, KirchhoffNet relies on Kirchhoff’s current law (KCL) to propagate information, which distinguishes itself from traditional DNNs where discrete layers (e.g., linear, convolution, and attention layers) are stacked in sequence. Moreover, we empirically show that even without any such traditional layers, KirchhoffNet can achieve state-of-the-art performances on a diverse range of complex machine learning problems, including three regression, seven generation, two density matching, and three image classification tasks. Furthermore, as a KirchhoffNet has the potential to be implemented using an analog integrated circuit, its on-chip forward calculation consistently can complete within a short timeframe, irrespective of the number of parameters. This appealing property positions KirchhoffNet as a promising lower-power analog computing paradigm for large-scale DNNs.
\section{Proposed Method}\label{sec:method}
\subsection{KirchhoffNet Definition --- KCL-based Information Flow}\label{sec:kcl}

KirchhoffNet or a circuit\footnote{In this paper, we will interchangeably use these two terms.} is best described by a directed graph $\mathcal{G}=(\mathcal{V},\mathcal{E})$, where $\mathcal{V}=\{n_i\}_{i=0}^N$ represents $(N+1)$ nodes, and $\mathcal{E}$ specifies directed edges among nodes. The $i$-th node is associated with a scalar \emph{nodal voltage} $v_i\in\mathbb{R}$, where $i\in\{0,1,\cdots,N\}$. Notably, we emphasize that the node $n_0$ is the \emph{ground node} so its associated $v_0$ is fixed to zero, following the convention of circuit theory~\cite{desoer1969basic,van1974network}. For later simplicity, we denote $\mathbf{v}=[v_1,v_2,\cdots,v_N]^T\in\mathbb{R}^N$. If the set $\mathcal{E}$ contains an edge from a source node $n_s$ to a destination node $n_d$, then a \emph{branch current} $i_{sd}\in \mathbb{R}$ will be generated flowing from $n_s$ to $n_d$. Now, applying KCL to some node $n_j$ (where $j=1,2,\cdots,N$), we obtain:
\begin{equation}\label{eq:KCL}
    \sum_{n_s\in\mathcal{N}^{s}(n_j)}\, i_{sj}=\sum_{n_d\in\mathcal{N}^{d}(n_j)}\, i_{jd}
\end{equation}
where $\mathcal{N}^{d}(n_j)=\{n_d\,|\,n_j\rightarrow n_d\in\mathcal{E}\}$ represents all destination nodes connected to $n_j$. Similarly, $\mathcal{N}^{s}(n_j)=\{n_s\,|\,n_s\rightarrow n_j\in\mathcal{E}\}$ represents all source nodes connected to $n_j$. Namely, KCL states that the sum of branch currents flowing into $n_j$ represented by the left-hand side of Eq.~(\ref{eq:KCL}) equals the sum of branch currents flowing out of $n_j$ represented by the right-hand side of Eq.~(\ref{eq:KCL}).

The branch current $i_{sd}$ is generated by a \emph{device} connecting from $n_s$ to $n_d$, and the value of $i_{sd}$ is related to the device parameter $\theta$ and the node voltages $\{v_s,v_d\}$. As illustrative examples, when a current source, a conductance, or a capacitance is connected from $n_s$ to $n_d$, we have the following expressions for $i_{sd}$, respectively: 
\begin{equation}\label{eq:source_branch}
    \begin{aligned}
        \text{Source branch:}\quad    i_{sd}&=\theta \\
    \end{aligned}    
\end{equation}
\begin{equation}\label{eq:cond_branch}
    \begin{aligned}
        \text{Conductive branch:}\quad    i_{sd}&=\theta (v_{s}-v_{d})\\
    \end{aligned}    
\end{equation}
\begin{equation}\label{eq:cap_branch}
    \begin{aligned}
        \text{Capacitive branch:}\quad      i_{sd}&=\theta (\dot{v}_s-\dot{v}_d)\\
    \end{aligned}    
\end{equation}
where we use $\dot{v}=dv/dt$ to represent the derivative of a node voltage $v$ with respect to the time variable $t$. As has been studied comprehensively in elementary circuit theory~\cite{desoer1969basic,van1974network}, a KirchhoffNet $\mathcal{G}$ with current-voltage relations shown in Eqs.~(\ref{eq:source_branch})-(\ref{eq:cap_branch}) and governed by KCL shown in Eq.~(\ref{eq:KCL}) can be solved by a set of ordinary differential equations (ODEs):
\begin{equation}\label{eq:basic_rc_circuit}
    \mathbf{C}\dot{\mathbf{v}}+\mathbf{G}\mathbf{v}=\mathbf{b}
\end{equation}
where $\{\mathbf{C}\in\mathbb{R}^{N\times N},\mathbf{G}\in\mathbb{R}^{N\times N},\mathbf{b}\in\mathbb{R}^N\}$ are some constants dependent on all parameters $\theta$ on all edges (devices), and $\mathbf{v}$ are the unknown node voltages required to be solved. At this point, we make three important comments: (i)~such a system is called an RC circuit and has been well studied, and is straightforward to realize in hardware. (ii)~Eq.~(\ref{eq:basic_rc_circuit}) naturally coincides with the formulation of continuous-depth neural networks (i.e., Neural ODE)~\cite{chen2018neural,dupont2019augmented}, if we view the node voltage vector $\mathbf{v}$ as the neural network feature variable. (iii)~An RC circuit alone is insufficiently general to work as a neural network, because the function governing the ODE dynamics is linear in $\mathbf{v}$, and thus the change of $\mathbf{v}$ along $t$ is restricted.

\begin{figure}[!htb]
    \centering
    \includegraphics[width=1.0\linewidth]{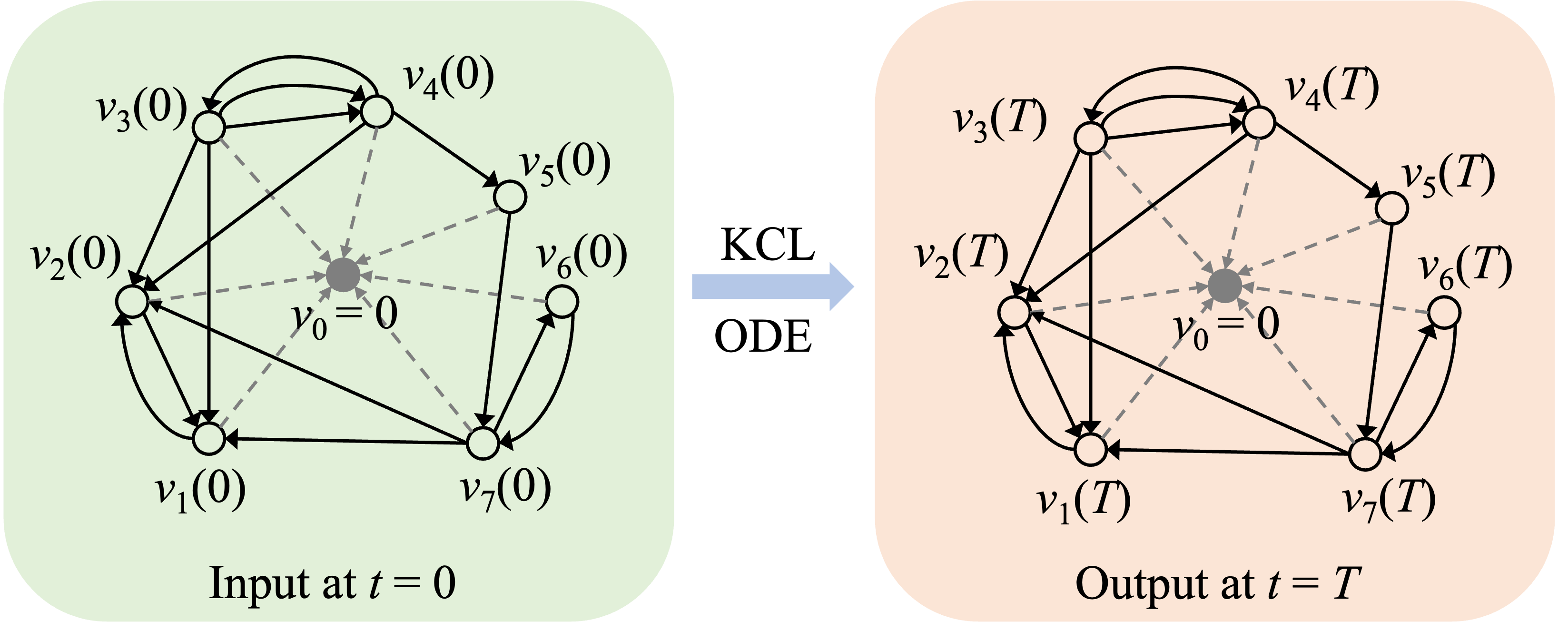}
    \vskip -0.1in
    \caption{An example of KirchhoffNet with $N=7$. The dashed grey lines represent fixed capacitive branches, while the solid black lines represent learnable non-linear branches. The node voltages $\mathbf{v}$ at $t=0$ and $t=T$ are respectively taken as the input and the output.}
    \label{fig:schematic}
\end{figure}

Now, to improve the representation capability, we deliberately introduce non-linear current-voltage relations and design our KirchhoffNet as shown in Fig.~\ref{fig:schematic}. Specifically, we enforce that all nodes $\{n_i\}_{i=1}^N$ connect to the ground node $n_0$ via a capacitance shown in Eq.~(\ref{eq:cap_branch}) with parameter value $\theta$. We emphasize that there will be $N$ such capacitive devices in total, and all their parameter values $\theta$ are identical, not learnable, and always fixed in training. Then, for the rest of the edges $\{n_s \rightarrow n_d\}$ specified by $\mathcal{E}$, their parameters are all learnable and have a non-linear current-voltage relation: $i_{sd}=g(v_s,v_d,\boldsymbol{\theta}_{sd})$, where $\boldsymbol{\theta}_{sd}$ represents a learnable parameter and $g(\cdot,\cdot,\cdot)$ is non-linear. In our paper, all these learnable edges are required to share the same function $g(\cdot,\cdot,\cdot)$, while their parameters might be different which will be determined by the training process. However, we clarify that it is not necessary for the same $g(\cdot,\cdot,\cdot)$ to be used universally and it is our choice here for implementation simplicity.

\begin{figure*}[!htb]
    \centering
    \includegraphics[width=0.95\linewidth]{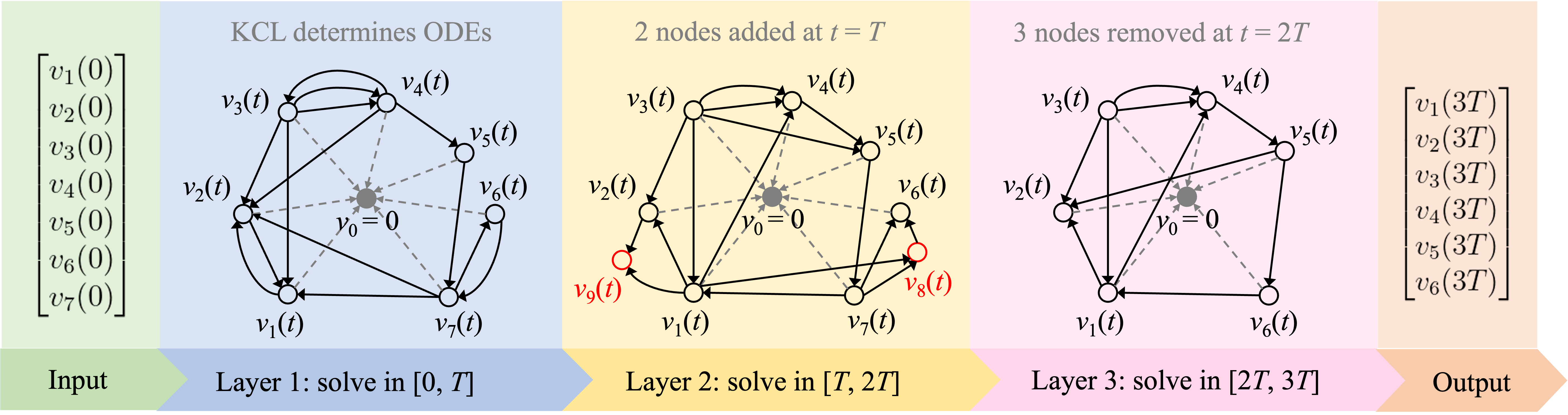}
    \caption{An example of a 3-layer KirchhoffNet with a 7-dimensional input and a 6-dimensional output. Note that at $t=T$ and $2T$, KirchhoffNet topology can change, such as the addition or removal of nodes and modifications to edge connections. In the case of new nodes being added, their associated node voltages are initialized to zero (e.g., $v_8(T)=v_9(T)=0$). The preceding Fig.~\ref{fig:schematic} depicts a 1-layer KirchhoffNet with a 7-dimensional input and output.}
    \label{fig:multi_layer_knet}
\end{figure*}

Explicitly writing out the impact of the fixed capacitance connecting from $n_j$ to $n_0$ in Eq.~(\ref{eq:KCL}) gives:
\begin{equation}\label{eq:special_KCL}
    \sum_{n_s\in\mathcal{N}^{s}(n_j)}\, i_{sj}\,=\,\theta\dot{v}_j\,+\, \sum_{n_d\in\mathcal{N}^{d}(n_j)}\, i_{jd}\; .
\end{equation}
Note that we have slightly abused the notation of $\mathcal{N}^{d}(n_j)$ in Eq.~(\ref{eq:special_KCL}).\footnote{Strictly, $\mathcal{N}^{d}(n_j)$ in Eq.~(\ref{eq:special_KCL}) is defined as $\{n_d\,|\,n_j\rightarrow n_d$ is not the capacitive branch and in $\mathcal{E}\}$.} The above equation can be further simplified to:
\begin{equation}\label{eq:dynamics}
\begin{aligned}
        \theta\dot{v}_j 
        &=\sum_{s\in\mathcal{N}^{s}(n_j)}\, i_{sj}-\sum_{\substack{d\in\mathcal{N}^{d}(n_j)}}\, i_{jd}\\
        &=  \sum_{s\in\mathcal{N}^{s}(n_j)}\, g(v_s,v_j,\boldsymbol{\theta}_{sj}) \\
        &\quad  - \sum_{\substack{d\in\mathcal{N}^{d}(n_j)}}\, g(v_j,v_d,\boldsymbol{\theta}_{jd}) \; .
\end{aligned}
\end{equation}
Similarly, we can derive ODEs for other node voltages, and overall a set of ODEs governs the evolution of $\mathbf{v}$. To clarify, multiple non-linear connections can exist between two nodes (see for example the edges between $n_3$ and $n_4$ in Fig.~\ref{fig:schematic}). The voltage values at $t=0$ and $t=T$, denoted as $\mathbf{v}(0)$ and $\mathbf{v}(T)$, are respectively the KirchhoffNet input and output~\cite{chen2018neural,dupont2019augmented}. Except for those $\theta$'s on the fixed capacitive branches connecting to the ground node $n_0$, all other $\boldsymbol{\theta}_{sd}$'s on the non-linear branches are learnable. The adjoint method~\cite{director1969generalized,chen2018neural,li2023adjoint} can be used to calculate the gradients of a customized loss function with respect to any learnable parameter $\boldsymbol{\theta}_{sd}$, and thus gradient-based optimizers (e.g., SGD and Adam) can be used for KirchhoffNet training. Notably, it has been validated that the adjoint method is more efficient than classical backward propagation in this regime~\cite{chen2018neural}.

When implementing a KirchhoffNet using a deep learning framework (e.g., Pytorch) in software, $T$ and $\theta$ are unit-less and purely numerical, which can be assigned values at our discretion. Following the literature~\cite{chen2018neural,dupont2019augmented}, it is common to set $T=\theta=1.0$. In contrast, a KirchhoffNet inherently resembles an analog circuit based on its definition and can be implemented in hardware. At that point, it becomes vital to consider physical units (e.g., seconds for time and Farads for capacitance). Further discussions on the hardware aspect are deferred to Section~\ref{sec:hardware}.

\subsection{KirchhoffNet Architecture --- Layer and Topology}\label{sec:multi_layer}

A modern DNN is usually constructed by stacking a few discrete layers (e.g., convolution, pooling, attention) in sequence and the number of these layers determines the model depth. However, as shown in Fig.~\ref{fig:schematic}, the presently defined KirchhoffNet does not adhere to the concept of layer. It has been established that making a neural network deeper can usually boost its performance~\cite{simonyan2014vgg}. This observation motivates us to define a deep KirchhoffNet as Fig.~\ref{fig:multi_layer_knet} demonstrates. Concretely, Fig.~\ref{fig:multi_layer_knet} presents an instance of a 3-layer KirchhoffNet, taking a 7-dimensional vector as input and producing a 6-dimensional vector as output. Its key is to use multiple time periods of duration $T$, where each one is defined to be one layer. Crucially, we emphasize that the topology, including nodes and edges, can vary in each layer (time period). For example, compared to the first layer, the second layer may include two additional nodes, $n_8$ and $n_9$, with node voltages $v_8(T)$ and $v_9(T)$ initialized to zero in Fig.~\ref{fig:multi_layer_knet}. The remaining node voltages, $v_1 \sim v_7$ at $t=T$ in the second layer can be initialized by inheriting their values from the first layer.

\begin{figure}[!ht]
        \centering
        \includegraphics[width=0.7\linewidth]{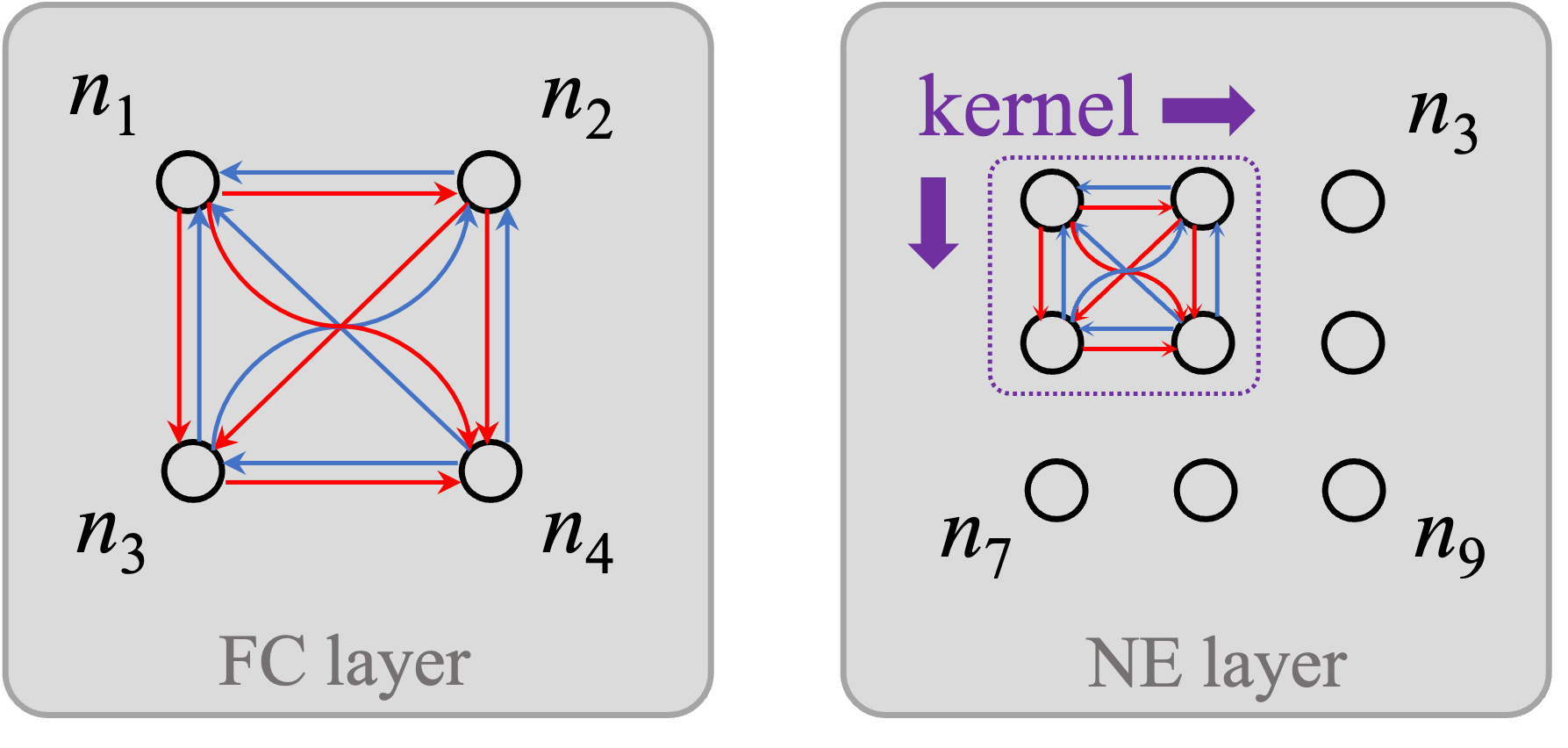}
        \vskip-4pt
    \caption{An illustration of a fully connected (FC) layer~{(left)} and a neighbor-emphasizing (NE) layer~{(right)} in a KirchhoffNet. Red edges denote connections from nodes with smaller indices to those with larger indices, while blue edges represent the opposite. In the right figure, a kernel slides across the given grid mesh, and for every possible kernel position, all nodes inside the kernel are fully connected. This results in edges between two nodes if they lie within a kernel. For instance, $n_1$ and $n_2$ are connected twice, while there is no connection between $n_1$ and $n_9$.}
    \label{fig:connect}
\end{figure}

In a conventional DNN, how we define each discrete layer (e.g., convolution or fully connected) impacts the overall performance of a DNN. Although there are no such types of discrete layers in a KirchhoffNet, we still need to define the topology in each time period, for the same reason that traditional DNNs have a variety of discrete layers. Fig.~\ref{fig:connect} showcases our proposed fully connected (FC) layer and the neighbor-emphasizing (NE) layer in a KirchhoffNet, serving as analogs to an FC layer and a convolution layer in a classical DNN. The NE layer contains more localized edges (e.g., no direct edges connecting $n_1$ and $n_9$) and is designed for image tasks. The right part of Fig.~\ref{fig:connect} uses a $3$-by-$3$ image and a kernel of size $1$ as an example, where the nine image pixel values are used to initialize $v_1 \sim v_9$ at $t=0$. Our definition of NE layer can be easily extended to the case of multiple-channel images (e.g., 3-channel RGB image). In practical applications, a projection (Proj) layer derived from the NE layer is commonly utilized. The Proj layer enhances an NE layer by incorporating additional ``projected" nodes. All original nodes in the NE layer establish full  connections with these projected nodes, while no connections exist among the projected nodes themselves.

The simplified code implementation of the proposed FC, NE, and Proj layers is presented in Fig.~\ref{code:pseudo_code}. It is essential to note that the code has undergone simplification, and certain details, such as enabling weight sharing among edges and introducing learnable connections to the ground node, have been omitted for the sake of clarity. Please refer to our open-source code base for more details.\footnote{Code will be released for full reproducibility of our experiments.}

\begin{figure}[!htb]
    \centering
\begin{tcolorbox}[
  colframe=gray,colback=white,boxrule=1pt,arc=4pt,
  boxsep=2pt,left=2pt,right=1pt,top=2pt,bottom=2pt]
\begin{minted}[fontsize=\small]{python}
import numpy as np
import torch
from torch.nn.functional import unfold

def fc_topo(num_node, repeat=1):
    edge = []
    for i in range(num_node):
        for j in range(num_node):
            if i != j:
                for _ in range(repeat):
                    edge.append([i, j])
    return edge


def ne_topo(c, w, h, k, repeat=1):
    
    n = c * w * h

    # Generate node indices.
    ind = torch.arange(1, 1 + n)
    ind = ind.view(c, w, h)

    # Unfold indices using
    # the given kernel size.
    
    # unsqueeze and squeeze for
    # utilizing the unfold function.
    ind = unfold(ind.unsqueeze(0).float(), k)
    ind = ind.squeeze(0)

    # Create edges
    edge = [
        [int(ind[j, i]), int(ind[l, i])]
        for i in range(ind.shape[1])
        for j in range(ind.shape[0])
        for l in range(ind.shape[0])
        if j != l
        for _ in range(repeat)
    ]
    return edge

def proj_topo(c, w, h, k, n, repeat=[1,1]):

    edge = ne_topo(c, w, h, k, repeat[0])
    m = np.max(np.array(edge))

    # add n projected nodes.
    for i in range(m + 1, m + 1 + n):
        for j in range(m + 1):
            for _ in range(repeat[1]):
                edge.append([i, j])
                edge.append([j, i])
    return edge

\end{minted}
\end{tcolorbox}
\vskip-4pt
    \caption{Simplified code implementations of a fully connected layer (FC layer), a neighbor emphasizing layer (NE layer), and a projection layer (Proj layer).}
    \label{code:pseudo_code}
\end{figure}

\subsection{Scalable Analog Computing}\label{sec:hardware}

The captivating aspect of KirchhoffNet is its direct connection to physical reality. Our formulation in Section~\ref{sec:multi_layer} essentially affirms that a KirchhoffNet can be realized as an analog circuit. In such physical hardware, we must account for physical units, typically represented in SI units, where time is measured in seconds (s), voltage in volts (V), current in amperes (A), and capacitance in farads (F).\footnote{For clarity, all symbols in our paper represent unit-less quantities. When necessary, we will explicitly specify the units associated with symbols by providing them immediately following the symbols.} Concretely, a hardware implemented KirchhoffNet could involve $N$ fixed capacitances, each with a value of 1~F, along with many learnable non-linear current-voltage devices. When we set the initial nodal voltage at $t=0$ s and let the circuit evolve to $t=DT$ s, the forward calculation is completed within this time frame. An example configuration entails $D=T=1$, but achieving a one-second forward run-time is not appealing at all.

We see that reducing the capacitance value can make the forward calculation much faster. To understand this mathematically, we take the simplest method for solving ODEs, the Forward Euler (FE) integration approximation, as an illustration. Essentially, in the forward calculation of KirchhoffNet, one needs to solve the ODE system shown in Eq.~(\ref{eq:dynamics}) from $t=0$ s to $t=DT$ s. FE accomplishes this by using numerical differentiation $\dot{v}\approx [v(t+\Delta t) - v(t)] / \Delta t$. Substituting into the left-hand side of Eq.~(\ref{eq:dynamics}) and simplifying the expression, we obtain:
\begin{equation}\label{eq:euler}
        v_j(t+\Delta t)= v_j(t) +\frac{\Delta t}{\theta} \left[\sum_{s\in\mathcal{N}^{s}(n_j)} i_{sj} - \sum_{\substack{d\in\mathcal{N}^{d}(n_j)}} i_{jd}\right] 
\end{equation}
where the terms inside the bracket are dependent on time $t$. Thus, when the circuit is solved at time $t$, Eq.~(\ref{eq:euler}) determines the circuit at time $(t+\Delta t)$. The step size $\Delta t$ usually must be very small for the solution to be accurate.


We notice that if we scale $\theta$ by a positive constant $a \ll 1$, i.e., $\theta \to a\times\theta$, then the value $v_j(r\Delta t)$ in the original case equals $v_j(ra\Delta t)$ in the scaled case for any $r\in\mathbb{N}$.\footnote{This equivalence can be directly proven by mathematical induction.} Namely, $v_j$ at $t$ originally will now equal the $v_j$ at $at$, i.e., $v_j(t)\to v_j(at)$. 
The equation below succinctly delineates the mapping of quantities from a unit-less KirchhoffNet in software (SW) to a hardware counterpart (HW), where considerations of physical units become imperative:
\begin{equation}\label{eq:sw_hw_map}
    \text{SW}\left\{\begin{array}{cc}
        D T \\
         \theta = 1.0
    \end{array}\right.\, \xrightarrow{\text{\scriptsize scaled by } a}\,
     \text{HW}\left\{\begin{array}{cc}
        D T\times a \text{ (s)}\\
         \theta = 1.0\times a \text{ (F)}
    \end{array}\right. \;.
\end{equation}
Our plain interpretation before, wherein $a=1$, yields a forward inference run-time at the second level. However, opting for $a=10^{-15}$ yields a forward run-time at the femtosecond level --- if we use $10^{-15}$ F capacitance and impose the initial nodal voltages on the circuit at $t=0$ fs, then it will finish the forward calculation at $t=DT$ fs. Most importantly, the above argument holds true \textbf{regardless of the number of KirchhoffNet parameters.} Alternatively, the effect of scaling by $a$ can be understood as simultaneously rescaling the units of time and capacitance.

Theoretically, the constant $a$ can be made arbitrarily small, indicating an extremely small capacitance, thereby achieving an arbitrarily small forward inference time. However, practical HW considerations impose limitations on the minimum attainable forward inference time. Various factors, such as the smallest implementable capacitance value, the minimum time interval between two voltage measurement operations, and the necessary minimum duration between two topology switching operations (as shown in Fig.~\ref{fig:multi_layer_knet}) for a deep KirchhoffNet, all establish lower bounds on the achievable forward inference time. 


\begin{figure}[!htb]
    \centering
    \vskip -0.1in
    \includegraphics[width=1.0\linewidth]{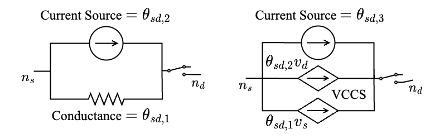}
    \vskip -0.1in
    \caption{{Left}: The schematic of a composite device made up of a conductance, a current source, and a one-sided switch, to realize the function $g$ shown in Eq.~(\ref{eq:non_linear_iv_relu2}). {Right}: Two voltage-controlled current sources (VCCSs), an independent current source, and a one-sided switch are needed to realize the function function $g$ shown in Eq.~(\ref{eq:non_linear_iv_relu3}). Note that the one-sided switch here is ideal, completely cutting off the current flowing from $n_d$ to $n_s$ while allowing the current to flow from $n_s$ to $n_d$.}
    \label{fig:device}
\end{figure}

Lastly, a critical consideration is the hardware implementation of the non-linear current-voltage relation $g(\cdot,\cdot,\cdot)$. We empirically find that the following four types of $g$ are sufficient for all of our numerical experiments in Section~\ref{sec:results}:
\begin{equation}\label{eq:non_linear_iv_relu2}
    g(v_s,v_d,\boldsymbol{\theta}_{sd})=\text{ReLU}(\theta_{sd,1}(v_s-v_d)+\theta_{sd,2})
\end{equation}
\begin{equation}\label{eq:non_linear_iv_tanh2}
    g(v_s,v_d,\boldsymbol{\theta}_{sd})=\tanh(\theta_{sd,1}(v_s-v_d)+\theta_{sd,2})
\end{equation}
\begin{equation}\label{eq:non_linear_iv_relu3}
    g(v_s,v_d,\boldsymbol{\theta}_{sd})=\text{ReLU}(\theta_{sd,1}v_s+\theta_{sd,2}v_d+\theta_{sd,3})
\end{equation}
\begin{equation}\label{eq:non_linear_iv_tanh3}
g(v_s,v_d,\boldsymbol{\theta}_{sd})=\tanh(\theta_{sd,1}v_s+\theta_{sd,2}v_d+\theta_{sd,3})
\end{equation}
where in Eqs.~(\ref{eq:non_linear_iv_relu2}) and~(\ref{eq:non_linear_iv_tanh2}), $\boldsymbol{\theta}_{sd}=[\theta_{sd,1},\theta_{sd,2}]^T$ contains two parameters, while in Eqs.~(\ref{eq:non_linear_iv_relu3}) and~(\ref{eq:non_linear_iv_tanh3}), $\boldsymbol{\theta}_{sd}=[\theta_{sd,1},\theta_{sd,2},\theta_{sd,3}]^T$ contains three. Let us take Eq.~(\ref{eq:non_linear_iv_relu2}) as an example. Comparing its form with Eqs.~(\ref{eq:source_branch})-(\ref{eq:cond_branch}), we realize that this nonlinear $g$ can be implemented in hardware by connecting a conductance and an independent current source in parallel, followed by connecting them in series to a one-sided switch, as shown in the left part of Fig.~\ref{fig:device}. Building upon this, Eq.~(\ref{eq:non_linear_iv_relu3}) introduces additional complexity, necessitating the inclusion of voltage-controlled current sources (VCCSs), as shown in the right part of Fig.~\ref{fig:device}. Similarly, Eqs.~(\ref{eq:non_linear_iv_tanh2}) and (\ref{eq:non_linear_iv_tanh3}) replace the ReLU function with $\tanh$, where we acknowledge the need for more detailed treatment; however, it is noteworthy that literature exists on realizing the hyperbolic function in the analog domain~\cite{Shakiba2021hperbolic}. In conclusion, the high-level schematic Fig.~\ref{fig:device} outlines a promising way for realizing KirchhoffNet with a physical circuit, but we note that substantial engineering and exploration of approximate implementations (e.g., using standard MOS transistors) will be needed. Our future work is dedicated to fabricating a KirchhoffNet in hardware, while the primary contribution of this paper resides in conceptualizing and numerically simulating KirchhoffNet.

\section{Numerical Results}\label{sec:results}

In this section, we justify the performance of the proposed KirchhoffNet on several machine learning tasks compared to state-of-the-art conventional DNNs. We re-emphasize that throughout all of our experiments, \textbf{KirchhoffNet does not have any traditional DNN layers}, but only those customized layers defined in Fig.~\ref{code:pseudo_code}. All of our experiments are implemented in Pytorch and run on a Linux cluster with V100 GPUs. For implementing the KirchhoffNet, we utilize the popular Torchdiffeq package~\cite{chen2018neural}. Note that the fast inference time at the femtosecond level mentioned in Section~\ref{sec:hardware} is promised on specialized hardware that we aim to fabricate in the future, not on GPUs.

On GPUs, ODE-based models are commonly observed to have slower execution times compared to conventional DNNs of equivalent parameters. We attribute this issue to the lack of a well-optimized ODE solver with the adjoint method enabled at the CUDA level. Given the current toolset, the present numerical experiments on ODE-based models are constrained in scale. For instance, to the best of our knowledge, classification results on the original ImageNet-22K using an ODE-based model has not been reported in any previous literature. The proposed KirchhoffNet, as an ODE-based model, grapples with this shared limitation. Thus, here we conduct numerical experiments \textbf{at a scale consistent with present ODE-based model literature, as used by}~\cite{chen2018neural,dupont2019augmented,xiao2023forward}.

\subsection{Regression}\label{sec:regression}

We conduct three regression experiments on the Friedman, California housing, and Diabete datasets, sourced from the Scikit-learn Python package. The state-of-the-art baseline model for these problems is multilayer perceptrons (MLPs). In this study, we use a KirchhoffNet containing one single FC layer with Eq.~(\ref{eq:non_linear_iv_relu2}) as the nonlinear current-voltage function, and a 4-layer MLP for comparison. To ensure fairness, all training procedures are executed with L2 loss, 400 epochs, a batch size of 64, a learning rate set to 0.0005, and the AdamW optimizer. For a comprehensive comparison, we design MLP and our KirchhoffNet in three different sizes: small, medium, and large. This involves varying the MLP hidden dimension and the number of nodes in the KirchhoffNet FC layer.

\begin{table}[!htb]
  \begin{threeparttable}
    \caption{Results of MLP and KirchhoffNet on Regression Problems}
    \label{tab:regression}
    \begin{tabular}{cccc}
    \toprule
          &  Friedman & Housing & Diabete \\ 
      \midrule
      MLP-s          & 9.97e-4 / 181 & 1.46e-2 / 391 & 2.71e-2 / 421 \\
      KirchhoffNet-s & 8.81e-4 / 168 & 1.55e-2 / 360 & 2.91e-2 / 528 \\
        \midrule
      MLP-m          & 1.35e-4 / 561 & 1.32e-2 / 841 & 2.61e-2 / 889 \\
      KirchhoffNet-m & 1.95e-5 / 528 & 1.25e-2 / 840 & 3.09e-2 / 840 \\
    \midrule
      MLP-l          & 6.28e-5 / 1596 & 1.27e-2 / 1551 & 2.67e-2 / 1581 \\
      KirchhoffNet-l & 1.11e-5 / 1520 & 1.16e-2 / 1520 & 3.48e-2  / 1520 \\
      \bottomrule
    \end{tabular}
    \begin{tablenotes}
      \item[1] Results are reported in the format `test loss / \#params' and averaged over 10 repeated runs.
    \end{tablenotes}
  \end{threeparttable}
\end{table}

\begin{figure}[!htb]
    \centering
    \includegraphics[width=1.0\linewidth]{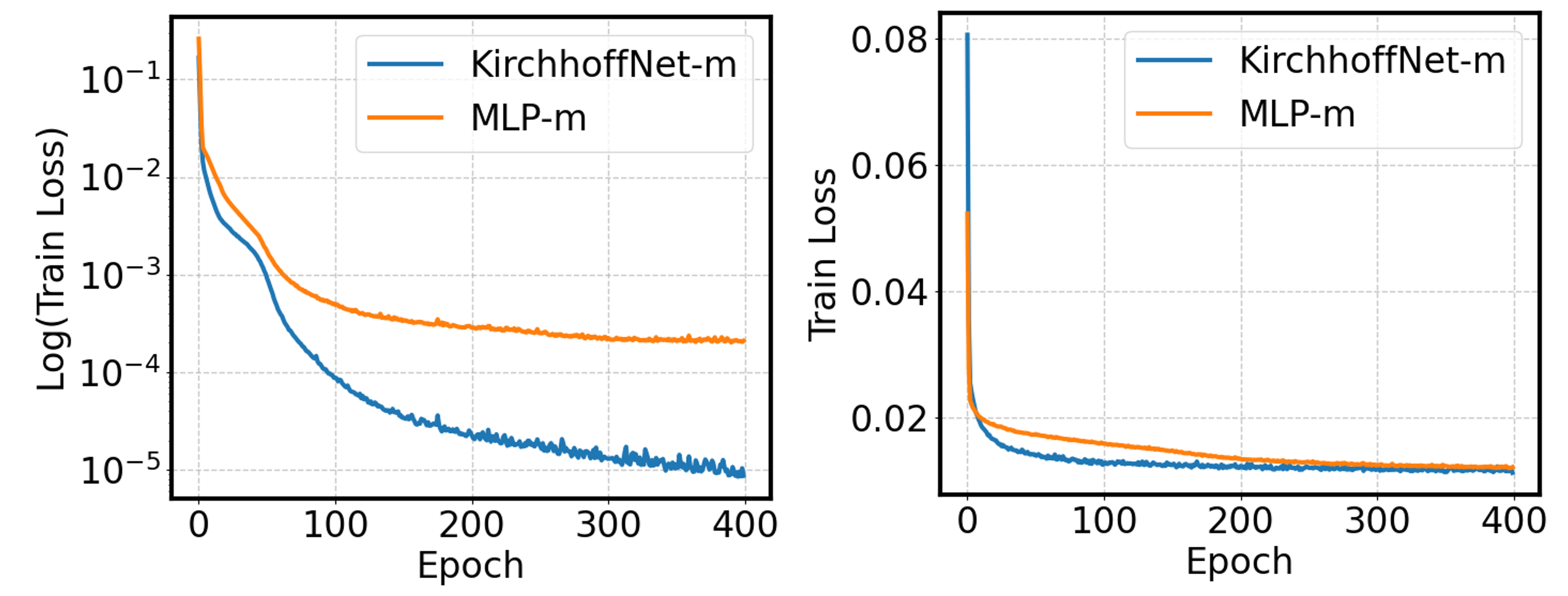}
    \caption{The raining loss is plotted against the number of epochs in a single run on the Friedman ({left}) and the Housing ({right}) dataset.}
    \label{fig:regress_train}
\end{figure}

Results are reported in Table~\ref{tab:regression} in the format of `test loss / \#params' after averaging over 10 repeated runs. We observe that when the models are of similar sizes, KirchhoffNet and MLP exhibit comparable L2 losses on unseen test samples. Occasionally, KirchhoffNet performs marginally better, and at other times, MLP outperforms. However, the differences in test loss metric values are negligible. Consequently, we assert that KirchhoffNet is comparable to an MLP with the same number of parameters in regression problems. Last but not least, two exemplary training loss curves are illustrated in Fig.~\ref{fig:regress_train}.

\subsection{Image Classification}

One popular deep learning problem is image classification. As our KirchhoffNet belongs to continuous-depth model and it will work as a specialized hardware accelerator for Neural ODE, the direct and fair comparison should be previous ODE-based models~\cite{chen2018neural,dupont2019augmented,massaroli2020dissecting}. For the MNIST classfication task, we use a KirchhoffNet with $D=2$ Proj layers and $T=1.0$. For the CIFAR-10 classification task, we use a KirchhoffNet with $D=4$ Proj layers and $T=0.5$. The training settings are similar, both with the AdamW optimizer and over $100$ epochs. Table~\ref{tab:image_classification} lists the test top-1 accuracies of different models on MNIST and CIFAR-10. Additionally, for reference, we report the performances of various graph neural networks (GNNs) and the classical ResNet in the bottom section. Notably, when available, results are presented in the format `mean$\pm$std' based on five repeated experiments. Analysis of Table~\ref{tab:image_classification} reveals that our model achieves state-of-the-art results compared to previous ODE-based models and outperforms GNNs. We also evaluate KirchhoffNet on the SVHN dataset, achieving approximately $92.1\%$ test top-1 accuracy, which surpasses Augmented Neural ODE~\cite{dupont2019augmented}. Additionally, we note that KirchhoffNet still exhibits a performance gap compared to convolutional neural networks (CNNs)~\cite{dwivedi2023benchmarking}, as CNNs are highly-tailored for image tasks.

\begin{table}[!htb]
    \centering
    \caption{Test Top-1 Accuracies (\%) of Different Models}
    \begin{tabular}{c|c|c|c}
    \toprule
    & Model & MNIST & CIFAR-10 \\
    \midrule
    Proposed & {KirchhoffNet} & $99.39\pm0.06$ & $73.41\pm0.12$ \\
    \midrule
    \multirowcell{4}{Compared \\ Baselines}
    & Neural ODE~\cite{dupont2019augmented} & $96.40\pm0.50$ & $53.70\pm0.20$ \\
    & Aug Neural ODE~\cite{dupont2019augmented} & $98.20\pm0.10$ & $60.60\pm0.40$ \\
    & IL Neural ODE~\cite{massaroli2020dissecting} & $99.10$ & $73.40$ \\
    & 2nd Neural ODE~\cite{massaroli2020dissecting} & $99.20$ & $72.80$ \\
    \midrule
    \multirowcell{8}{For \\ Reference}
    & MLP~\cite{dwivedi2023benchmarking} & $95.34\pm0.14$ & $56.34\pm 0.18$ \\
    & Vanilla GCN~\cite{dwivedi2023benchmarking} & $90.71\pm0.22$ & $55.71\pm0.38$ \\
    & GraphSage~\cite{dwivedi2023benchmarking} & $97.32\pm0.10$ & $65.77\pm0.31$ \\
    & GCN~\cite{dwivedi2023benchmarking} & $90.12\pm0.15$ & $54.14\pm0.39$ \\
    & MoNet~\cite{dwivedi2023benchmarking} & $90.81\pm0.03$ & $54.66\pm0.52$ \\
    & GAT~\cite{dwivedi2023benchmarking} & $95.54\pm0.21$ & $64.23\pm0.46$ \\
    & GatedGCN~\cite{dwivedi2023benchmarking} & $97.34\pm0.14$ & $67.31\pm0.31$ \\
    \bottomrule
    \end{tabular}
    \label{tab:image_classification}
\end{table}

\subsection{Generation and Density Matching}

\begin{figure*}[!ht]
    \centering
    \includegraphics[width=0.9\linewidth]{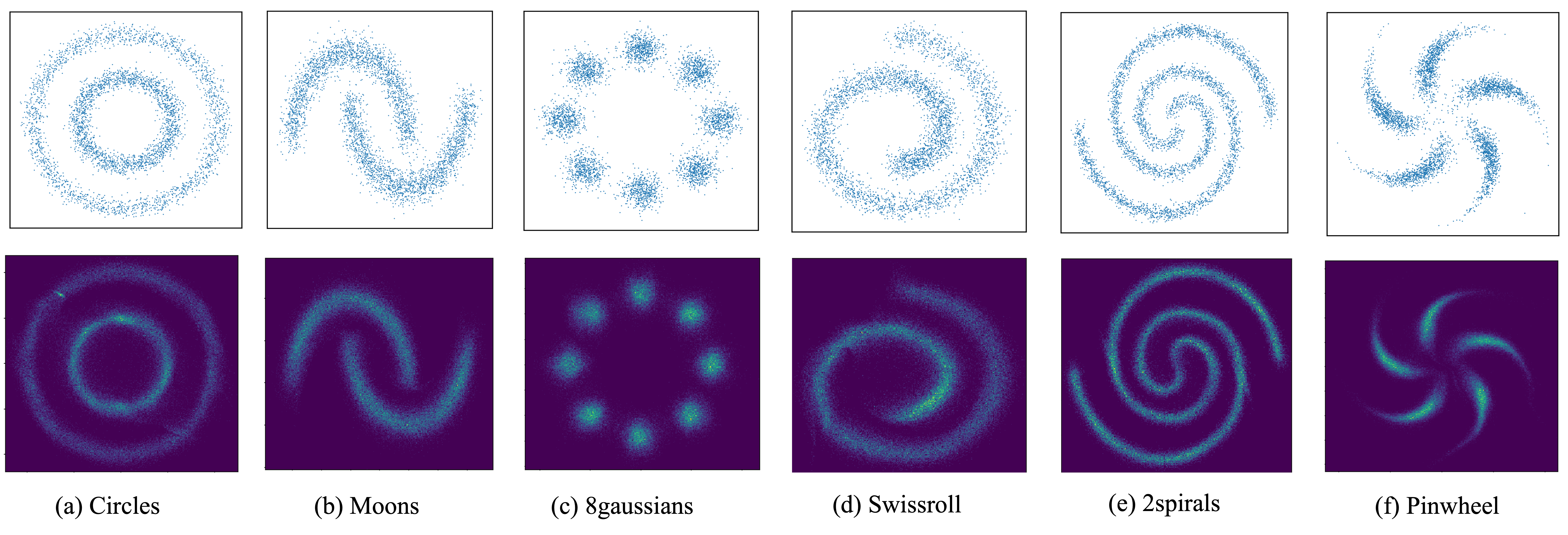}
    \vskip -6pt
    \caption{Visualizations of 2D generation testbenches. The top row showcases the provided training data samples, while the bottom row displays the generated samples produced by our KirchhoffNet after training. We observe a good alignment between our learned samples and the target in each testbench.}
    \label{fig:generation}
\end{figure*}

The instantaneous change of variable theorem in~\cite{chen2018neural} states that when we consider the variable $\mathbf{v}$ in Eq.~(\ref{eq:dynamics}) as a random variable, then the time derivative of the logarithm probability $\partial \log p(\mathbf{v}(t))/\partial t$ has an analytical formula related to the right-hand side of Eq.~(\ref{eq:dynamics}) in the ODE domain. Leveraging this, we can introduce probability into KirchhoffNet (i.e., a KirchhoffNet generating probabilistic outputs) and perform generation and density matching tasks. Further details about the definition of generation and density matching problems and their training losses can be found in the Appendix.



We first consider several 2D generation tasks used by~\cite{chen2018neural} and~\cite{dai2020learning}. In each generation task, a training dataset is provided, and we attempt to make the KirchhoffNet generate samples similar to the provided. The top row in Fig.~\ref{fig:generation} illustrates the provided training data in six tasks. We set the batch size to 512, the learning rate to $0.001$, the number of epochs to no less than $10000$, and employ an AdamW optimizer with a cosine learning rate scheduler. For the majority of tasks, we configure KirchhoffNet utilizing FC layers with parameters $D=4$ and $T=0.5$, employing Eq.~(\ref{eq:non_linear_iv_tanh3}) as the nonlinear current-voltage function. As part of an ablation study to demonstrate the robustness of our model and introduce some variations, we intentionally employ $D=2$ in the `Moons' testbench and $D=8$ in the `2spirals' testbench. It is noteworthy that while increasing $D$ from $4$ to $8$ in the `2spirals' testbench, we achieve improved results by reducing $T$ from $0.5$ to $0.2$.

\begin{figure*}[!htb]
    \centering
    \includegraphics[width=0.85\linewidth]{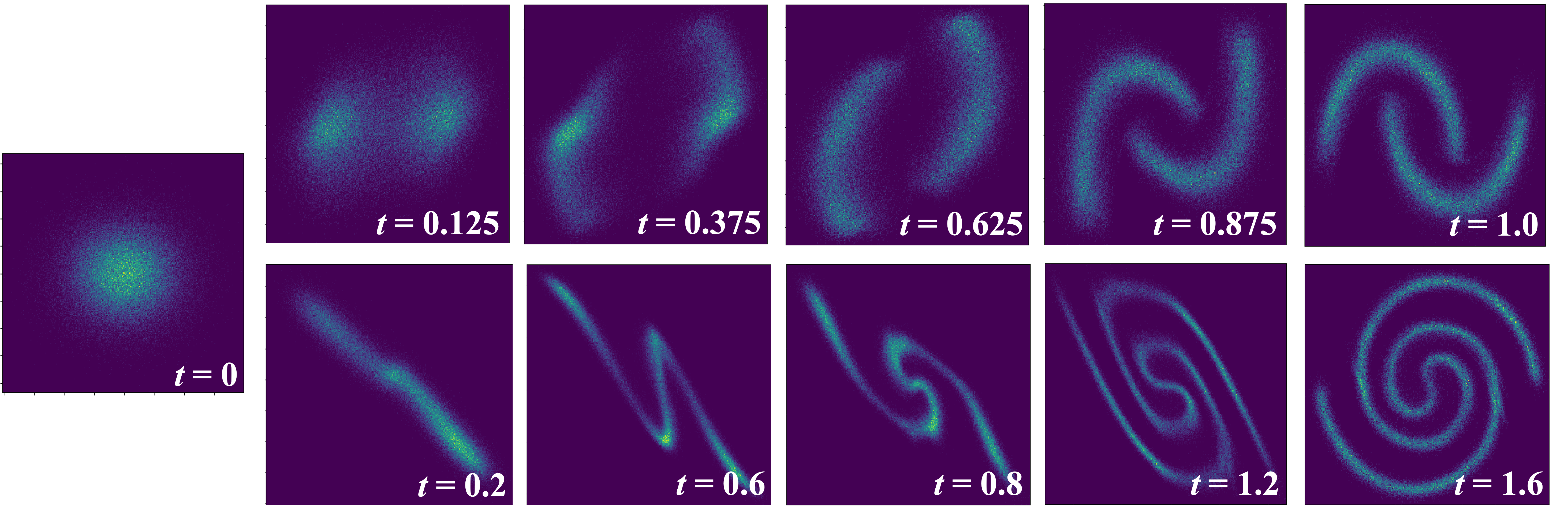}
    \vskip -6pt
    \caption{The intermediate generation results are showcased for the 'Moons' (top row) and the '2spirals' (bottom row) testbenches. It is important to note that in the 'Moons' example, we set $D=2$ and $T=0.5$, whereas in the '2spirals' example, we have $D=8$ and $T=0.2$.}
    \label{fig:generation_process}
\end{figure*}

The bottom row in Fig.~\ref{fig:generation} demonstrates the samples generated by the KirchhoffNet after training. We observe a good alignment between our learned samples and the target in each testbench. Furthermore, Fig.~\ref{fig:generation_process} presents the intermediate generation results for the 'Moons' and '2spirals' testbenches, clearly illustrating the smooth and intuitive learned data transformation in our KirchhoffNet. For conciseness, readers can refer to Figure 5 in~\cite{chen2018neural} for a comparison of our results with those generated by Neural ODE and Normalizing Flows.

\begin{figure}[!ht]
    \centering
    \includegraphics[width=0.65\linewidth]{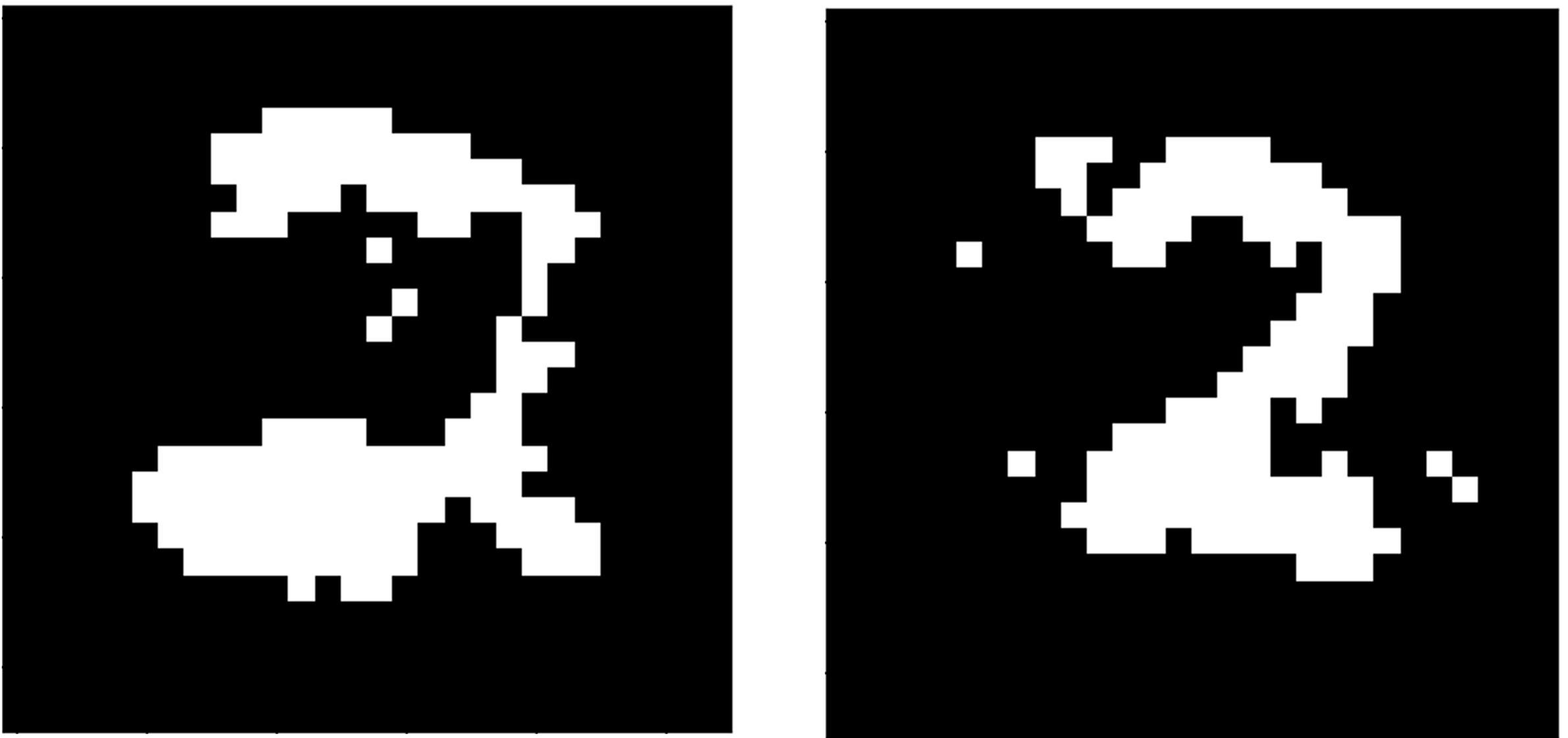}
    \vskip -6pt
    \caption{Two generated images of digit 2 by the KirchhoffNet are shown.}
    \label{fig:mnist_digit2}
\end{figure}

Next, we evaluate KirchhoffNet's performance in generating handwritten digits using MNIST as the training dataset. Due to the high simulation runtime cost, we must limit the size of our KirchhoffNet to be relatively small. Specifically, in this instance, we utilize a KirchhoffNet with $D=2$ NE layers and $T=1.0$, Eq.~(\ref{eq:non_linear_iv_tanh2}) as the current-voltage function, and train solely on a subset of digit 2 for 45 epochs. Two generated images are illustrated in Fig.~\ref{fig:mnist_digit2}. Despite minor imperfections, they are generally acceptable and resemble the digit 2.

\begin{figure}[!htb]
    \centering
    \includegraphics[width=0.6\linewidth]{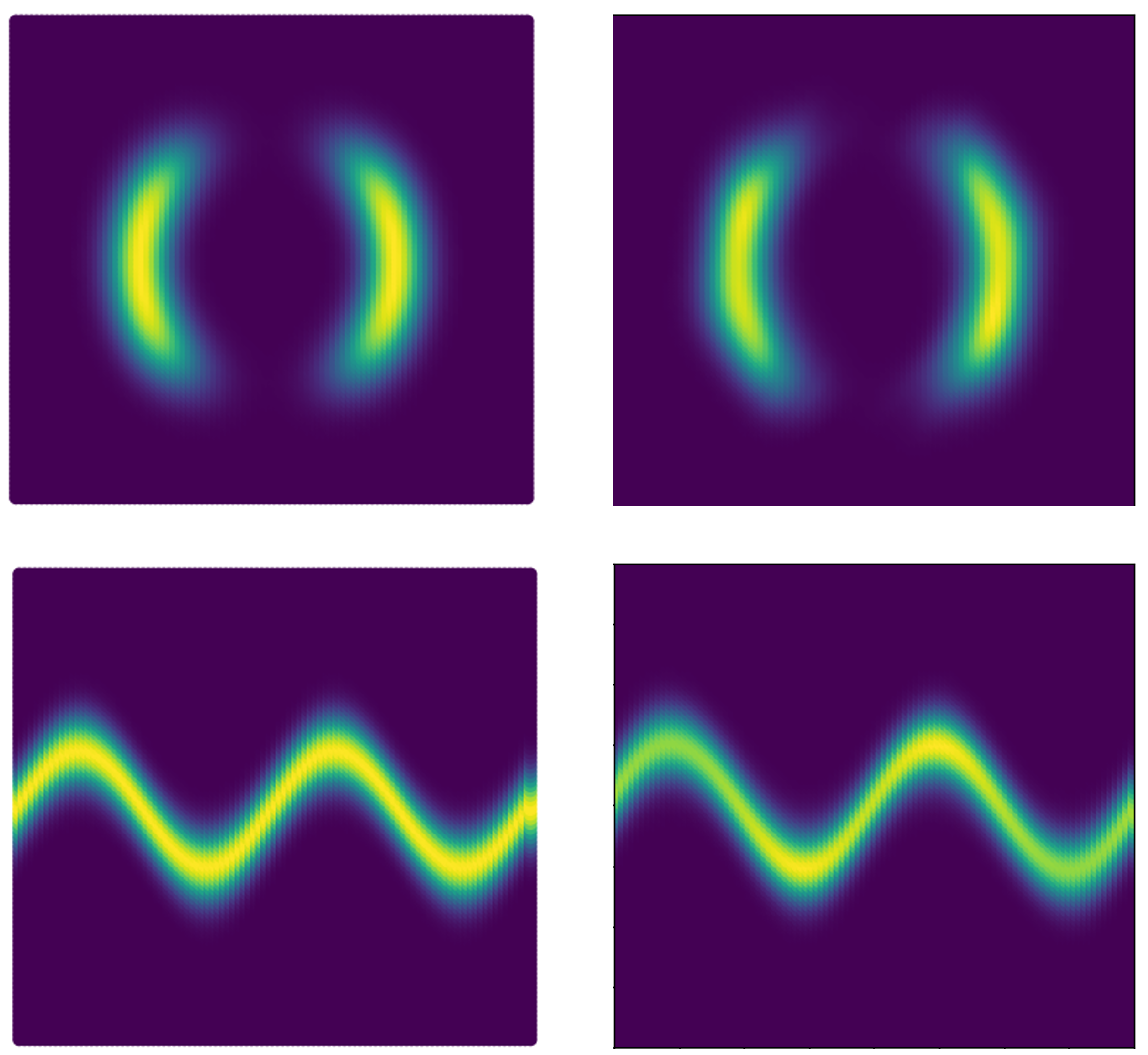}
    \vskip -6pt
    \caption{The target density function is displayed on the left, while the learned density function by the KirchhoffNet is shown on the right.}
    
    \label{fig:density_matching}
\end{figure}

Finally, we consider density matching tasks. In a density matching task, instead of a training dataset, we are provided a density function and attempt to learn the probability distribution it corresponds to. Here we set the batch size to 512, the learning rate to 0.001, the number of epochs to no less than $10000$, and employ an AdamW optimizer with a cosine learning rate scheduler. We use a KirchhoffNet with $D=8$ FC layers and $T=0.2$. As shown in Fig.~\ref{fig:density_matching}, KirchhoffNet can accurately learn the density function. For comparison with baselines, please refer to Fig.~4 in~\cite{chen2018neural}. 

\section{Discussion}\label{sec:discussion}

\subsection{Remarks on Models Related to KirchhoffNet}\label{sec:related_works}

From a mathematical perspective, KirchhoffNet is a specialized version of Neural ODE~\cite{chen2018neural}. Namely, the function governing the ODE dynamics has a concrete form in KirchhoffNet as shown in the right-hand side of Eq.~(\ref{eq:dynamics}), while Neural ODE directly uses a DNN. Notably, in Neural ODE, the variable $t$ lacks a physical interpretation; it merely serves as a parameter dictating when the ODE is integrated. Conversely, in KirchhoffNet, the variable $t$ encapsulates physical units, intrinsically tied to the on-chip forward run time within a real hardware setting. Alternatively, one can conceive of KirchhoffNet and its physical realization as a specialized hardware accelerator tailored for Neural ODE.

Analog computing, particularly its role as a deep learning accelerator, has garnered significant attention. An especially relevant precursor to our KirchhoffNet is the exploration of physically grounded analog neural networks in previous works~\cite{kendall2020training, kendall2021gradient, scellier2021deep}. In~\cite{kendall2021gradient}, the focus is primarily on physical systems characterized by an energy function, where training involves seeking equilibrium, manifested as the minimization of the energy function. The authors, for example, consider non-linear resistive networks and minimize the power to find the steady state (see Lemma 4.1 in~\cite{kendall2021gradient}).

It is important to emphasize that, despite KirchhoffNet updating node voltage values based on neighboring nodes, it diverges from the paradigm of message passing neural networks (MPNNs) and graph neural networks (GNNs). The difference lies in the fact that KirchhoffNet's parameters are situated on the edges, not nodes as with MPNNs or GNNs. For instance, employing Eq.~(\ref{eq:non_linear_iv_relu2}) as the i-v function implies that a node $n_j$ with $N$ neighboring nodes will be governed by $2N$ parameters influencing its evolution over time, as illustrated in Eq.~(\ref{eq:dynamics}). Namely, the number of parameters governing node updates is contingent upon the count of nodes within its vicinity. This stands in contrast to MPNNs or GNNs, where the same number of parameters always govern node updates. Finally, it is worth mention that having learnable weights on edges in KirchhoffNet shares similar idea with the emerging KAN net~\cite{liu2024kan}.

\subsection{Limitations and Future Works}\label{sec:future}

In this paper, our key contribution is the conceptualization of KirchhoffNet and the initial demonstration of its capabilities by simulating the analog circuit. In its infancy, there is much to investigate and improve:

\begin{itemize}[left=0pt]
    \item \textbf{Large-scale experiments}: Our current experiments, aligned with existing ODE-based model literature~\cite{chen2018neural,dupont2019augmented,massaroli2020dissecting}, are limited in scale. Extending KirchhoffNet and other ODE-based models to larger-scale experiments is currently unfeasible.
    \item \textbf{Architecture design}: Proposed KirchhoffNet layers include FC, FE, and Proj. Further investigation into model architecture is vital, as our current performance matches ODE-based models but falls short of CNNs in image classification.
    \item \textbf{Circuit-level verification}: Python and PyTorch drive our paper's numerical evaluations, and thus, overlooking non-ideal behaviors and real-world considerations is inevitable. Simulating KirchhoffNet in SPICE-derivative circuit-level simulators is a crucial future step.
    \item \textbf{Hardware fabrication}: Building on the previous point, hardware design and fabrication will enable further exploration of the efficiency and effectiveness of KirchhoffNet.
\end{itemize}

\section{Conclusions}\label{sec:conclusion}

In this paper, we introduce a novel class of analog neural networks named KirchhoffNet, grounded in Kirchhoff's  laws. KirchhoffNet is conceptualized as an analog circuit operating in the time domain, utilizing initial node voltages as inputs and node voltages at specific time points as outputs. Our empirical results demonstrate its capability to achieve state-of-the-art performance across diverse deep-learning problems. Furthermore, KirchhoffNet has the potential to be  realized as an analog integrated circuit in hardware, enabling short on-chip forward run times and potentially low power consumption. This  appealing property fuels our optimism regarding KirchhoffNet's potential to contribute to the landscape of analog neural networks.

\appendix
\label{sec:appendix_training_loss}

A generation problem is usually performed by a neural network that can generate probabilistic outputs (e.g., GAN, VAE, flow-based model, diffusion model). Specifically, the neural network is denoted as $q_{\boldsymbol{\theta}}(\mathbf{x})$, where $\boldsymbol{\theta}$ represents the learnable parameters of the network, and $q_{\boldsymbol{\theta}}(\cdot)$ signifies a distribution from which we can readily draw a sample $\mathbf{x}$. The objective is to ensure that the sampled $\mathbf{x}$ closely resembles the examples provided in the training dataset $\mathcal{D}=\{\mathbf{x}_i\}_{i=1}^N$. This alignment is accomplished by minimizing the negative logarithm likelihood with respect to $\boldsymbol{\theta}$. Importantly, this training loss is derived based on:
\begin{equation}\label{eq:loss_generation}
\begin{aligned}
    \text{KL}[p(\mathbf{x})||q_{\boldsymbol{\theta}}(\mathbf{x})]&=\mathbb{E}_p[\log\frac{p(\mathbf{x})}{q_{\boldsymbol{\theta}}(\mathbf{x})}]\approx \frac{1}{N}\sum_{i=1}^N \log\frac{p(\mathbf{x}_i)}{q_{\boldsymbol{\theta}}(\mathbf{x}_i)}\\
    &\propto-\frac{1}{N}\sum_{i=1}^N\log q_{\boldsymbol{\theta}}(\mathbf{x}_i)
\end{aligned}
\end{equation}
where $p(\cdot)$ denotes the unknown data-generating distribution from which the training dataset is sampled, i.e., $\mathbf{x}_i$ is an i.i.d. sample drawn from $p(\cdot)$. Note that the last line only keeps the terms dependent on $\boldsymbol{\theta}$.

Alternatively, in a density matching problem, there are no training samples, but a black-box density function $u(\mathbf{x})$ is provided. The function $u(\mathbf{x})$ corresponds a distribution $p(\mathbf{x})$ upon an unknown normalization constant $Z$, i.e., $p(\mathbf{x})=Z u(\mathbf{x})$, and our goal is to make $q_{\boldsymbol{\theta}}(\mathbf{x})$ to mimic $p(\mathbf{x})$. This can be done by minimizing the KL divergence loss:
\begin{equation}
\begin{aligned}
        \text{KL}[q_{\boldsymbol{\theta}}(x) || p(\mathbf{x})]&=\mathbb{E}_{q_{\boldsymbol{\theta}}}[\log\frac{p(\mathbf{x})}{q_{\boldsymbol{\theta}}(\mathbf{x})}]\propto\mathbb{E}_{q_{\boldsymbol{\theta}}}[\log\frac{u(\mathbf{x})}{q_{\boldsymbol{\theta}}(\mathbf{x})}]\\
        &\approx \frac{1}{N}\sum_{i=1}^N \log\frac{u(\mathbf{x}_i)}{q_{\boldsymbol{\theta}}(\mathbf{x}_i)}\quad \mathbf{x}_i\sim q_{\boldsymbol{\theta}}(\cdot)
\end{aligned}
\end{equation}
where $\mathbf{x}_i$ in the second line depends on $\boldsymbol{\theta}$, as it is sampled from $q_{\boldsymbol{\theta}}$, and cannot be omitted, which differs from Eq.~(\ref{eq:loss_generation}). It is important to highlight that the setups, training methodologies, and applications of generation and density matching are fundamentally distinct. In the case of our KirchhoffNet, its output $\mathbf{v}(DT)$ is probabilistic and serves the role equivalent to the variable $\mathbf{x}$. Thus, sampling an $\mathbf{x}$ from $q_{\boldsymbol{\theta}}(\cdot)$ in our context is akin to initially sampling a value from a standard Gaussian distribution, taking it as $\mathbf{v}(0)$, and subsequently returning the value $\mathbf{v}(DT)$ as $\mathbf{x}$. Please refer to~\cite{chen2018neural} for more details.

\bibliography{sample}
\bibliographystyle{IEEEtran}

\end{document}